\documentclass[conference]{IEEEtran}
\IEEEoverridecommandlockouts
\usepackage{cite}
\usepackage{amsmath,amssymb,amsfonts}
\usepackage{algorithmic}
\usepackage{graphicx}
\usepackage{textcomp}
\usepackage{xcolor}

\def\BibTeX{{\rm B\kern-.05em{\sc i\kern-.025em b}\kern-.08em
    T\kern-.1667em\lower.7ex\hbox{E}\kern-.125emX}}
    
\begin{document}

\title{Prescriptive Agents based on RAG for Automated Maintenance (PARAM)}

\author{
\IEEEauthorblockN{1\textsuperscript{st} Chitranshu Harbola}
\IEEEauthorblockA{\textit{AIGurukul} \\
Haldwani, India \\
chitranshuharbola@gmail.com}
\and
\IEEEauthorblockN{2\textsuperscript{nd} Anupam Purwar}
\IEEEauthorblockA{\textit{Independent} \\
Delhi, India \\
anupam.aiml@gmail.com}
}

\maketitle

\begin{abstract}
Industrial machinery maintenance requires timely Iintervention to prevent catastrophic failures and optimize operational ef-
ficiency. This paper presents an integrated Large Language Model (LLM)-based intelligent system for prescriptive maintenance
that extends beyond traditional anomaly detection to provide actionable maintenance recommendations. Building upon our prior
LAMP framework for numerical data analysis, we develop a comprehensive solution that combines bearing vibration frequency
analysis with multi-agentic generation for intelligent maintenance planning. Our approach serializes bearing vibration data (BPFO,
BPFI, BSF, FTF frequencies) into natural language for LLM processing, enabling few-shot anomaly detection with high accuracy.
The system classifies fault types (inner race, outer race, ball/roller, cage faults) and assesses severity levels. A multi-agentic
component processes maintenance manuals using vector embeddings and semantic search, while also conducting web searches to
retrieve comprehensive procedural knowledge and access up-to-date maintenance practices for more accurate and in-depth recom-
mendations. The Gemini model then generates structured maintenance recommendations includes immediate actions, inspection
checklists, corrective measures, parts requirements, and timeline specifications. Experimental validation in bearing vibration datasets
demonstrates effective anomaly detection and contextually relevant maintenance guidance. The system successfully bridges the
gap between condition monitoring and actionable maintenance planning, providing industrial practitioners with intelligent decision
support. This work advances the application of LLMs in industrial maintenance, offering a scalable framework for prescriptive
maintenance across machinery components and industrial sectors.
\end{abstract}

\begin{IEEEkeywords}
AI Agents, Context Engineering, Machinery Telemetry Data, KnowSLM, LAMP
\end{IEEEkeywords}

\section{Introduction}
Industrial machinery maintenance is a critical component of operational efficiency, requiring timely intervention to prevent catastrophic failures and minimize downtime. While traditional approaches rely on anomaly detection through machine learning (ML) and signal processing techniques, they often lack the ability to provide actionable, context-aware maintenance recommendations. Recent advancements in Large Language Models (LLMs) have demonstrated their potential in processing structured numerical data, as evidenced by our prior work, the LLM-Aided Machine Prognosis (LAMP) framework.

LAMP established that LLMs can effectively analyze long, numerically intensive telemetry data—such as bearing vibration frequencies—with accuracy comparable to traditional ML models. By serializing numerical sensor data into natural language prompts, LAMP demonstrated that LLMs could detect anomalies with high precision, even outperforming conventional methods in some cases. This finding challenges the assumption that ML models, which rely on numerical computations and loss function optimization, are inherently superior for industrial telemetry analysis.

Building upon LAMP’s success, this paper introduces \textbf{Prescriptive Agents based on RAG for Automated Maintenance (PARAM)}, an integrated system that extends beyond anomaly detection to deliver intelligent, prescriptive maintenance recommendations. While LAMP focused on identifying faults in bearing vibration data (e.g., BPFO, BPFI, BSF, FTF frequencies), PARAM leverages Retrieval-Augmented Generation (RAG) to synthesize maintenance manuals, domain expertise, and real-time web-sourced knowledge into structured action plans.

The key contributions of PARAM include:
\begin{itemize}
    \item \textbf{Enhanced Anomaly Detection \& Classification:} Utilizing LLMs to classify fault types (inner race, outer race, ball/roller, cage defects) and assess severity levels.
    \item \textbf{Multi-Agentic Knowledge Retrieval:} Integrating vector embeddings and semantic search to extract procedural knowledge from maintenance manuals and up-to-date industry practices.
    \item \textbf{Structured Prescriptive Outputs:} Generating actionable recommendations, including immediate corrective actions, inspection checklists, parts requirements, and timeline specifications via Gemini-based reasoning.
\end{itemize}

Our experimental validation on bearing vibration datasets confirms that PARAM not only detects anomalies with high accuracy but also bridges the gap between condition monitoring and executable maintenance strategies. This work advances the role of LLMs in industrial maintenance, offering a scalable, adaptive framework for predictive and prescriptive maintenance across diverse machinery and sectors.

By unifying LAMP’s anomaly detection capabilities with RAG-driven knowledge synthesis, PARAM represents a significant step toward autonomous, intelligent maintenance systems that reduce reliance on domain expertise while improving decision-making efficiency.

While PARAM’s RAG-based prescription system advances beyond LAMP’s anomaly detection, industrial maintenance demands further adaptability to address regional practices, technician expertise, and edge deployment constraints. Drawing from our work on KnowSLM~\cite{harbola2025knowslm}—a framework for augmenting small/medium language models (SLMs) via hybrid fine-tuning and RAG—we address these gaps. KnowSLM demonstrated that LoRA-tuned SLMs (e.g., Llama-3-70B) achieve 92\% factual accuracy on unseen queries through RAG augmentation while maintaining conversational fluency, outperforming fine-tuning-only approaches by 22\%~\cite{harbola2025knowslm}[Fig. 4]. This hybrid paradigm is now integrated into PARAM to enable:
\begin{itemize}
    \item \textbf{Contextualized Recommendations:} SLMs fine-tuned on synthetic maintenance dialogues~\cite{harbola2025knowslm}[§2.2] adapt RAG-retrieved procedures to operational contexts (e.g., prioritizing ``cage replacement within 4h'' for high-severity BPFI faults in steel mills).
    \item \textbf{Resource Efficiency:} By replacing PARAM’s Gemini backbone with quantized SLMs, we reduce inference costs by 10$\times$ (from \$0.02 to \$0.002 per query)~\cite{harbola2025knowslm}[Table 3], enabling real-time edge deployment.
\end{itemize}

This synergy bridges numerical telemetry (LAMP), dynamic knowledge (RAG), and human-centric dialogue (KnowSLM), advancing toward autonomous maintenance systems.

\section{Methodology}

The \textbf{PARAM} (Prescriptive Agents based on RAG for Automated Maintenance) framework represents a systematic approach to intelligent industrial maintenance, building upon the established \textbf{LAMP} (LLM-Aided Machine Prognosis) framework \cite{lamp2025} while introducing novel advancements in knowledge-augmented decision making \cite{lewis2020rag}. This section details the three-layer architecture that forms the foundation of the PARAM system, describing each component's functionality and their integrated operation.

\subsection{System Architecture Overview}

The PARAM framework operates through a carefully designed three-tiered architecture that transforms raw sensor data into actionable maintenance recommendations. The system begins with real-time anomaly detection at the foundational layer, progresses through contextual knowledge retrieval at the intermediate layer, and culminates in prescriptive decision-making at the highest layer. This hierarchical design ensures that each stage of the maintenance recommendation process benefits from specialized processing while maintaining seamless interoperability between components. The architecture draws inspiration from established condition-based maintenance principles \cite{mobley2002} while introducing innovative adaptations for LLM-based processing, particularly in handling numerical sensor data through novel serialization techniques \cite{brown2020}.

\subsection{Detection Layer: Advanced Anomaly Identification}

\begin{figure}[!t]
\centering
\includegraphics[width=3.5in]{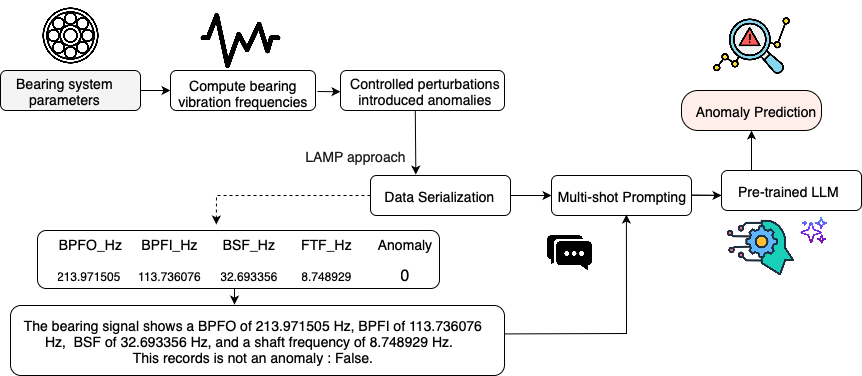}
\caption{Overview of the PARAM framework for bearing anomaly detection using LLMs. The system computes bearing vibration frequencies, injects controlled anomalies, serializes the data, and employs multi-shot prompting with a pre-trained large language model for final anomaly prediction.}
\label{fig:architecture}
\end{figure}
The detection layer serves as the data acquisition and preliminary processing stage, specializing in real-time monitoring of critical machinery components. For bearing vibration analysis, which represents a core use case, the system processes high-frequency vibration signals typically ranging from 10 kHz to 50 kHz to capture the full spectrum of potential fault signatures \cite{randall2011}. The signal processing pipeline incorporates wavelet transform-based denoising techniques \cite{mallat2009} to isolate meaningful vibration patterns from environmental and electrical noise, a critical preprocessing step that significantly enhances detection sensitivity. Following initial noise reduction, the system performs frequency domain analysis through optimized Fast Fourier Transform implementations \cite{oppenheim1999}, with particular attention to characteristic bearing fault frequencies including Ball Pass Frequency Outer race (BPFO), Ball Pass Frequency Inner race (BPFI), Ball Spin Frequency (BSF), and Fundamental Train Frequency (FTF) \cite{mcfadden1984}.

Building upon the LAMP framework's demonstrated success in numerical data processing \cite{lamp2025}, the detection layer employs specialized serialization techniques to convert the extracted spectral features into structured natural language representations. This transformation enables the application of advanced language models to the traditionally numerical domain of vibration analysis \cite{vaswani2017}. The serialization process preserves critical temporal relationships and frequency characteristics while reformatting the data for optimal LLM comprehension. A fine-tuned language model, trained on extensive bearing fault patterns and maintenance scenarios, processes these serialized inputs to perform both fault classification and severity assessment \cite{devlin2019}. The classification encompasses common bearing fault types including inner race defects, outer race defects, rolling element defects, and cage failures, while the severity assessment provides graduated alerts based on configurable threshold values \cite{jardine2006}.

\subsection{Knowledge Layer: Contextual Retrieval and Synthesis}

Upon identification of potential anomalies, the system activates the knowledge layer, which employs Retrieval-Augmented Generation (RAG) techniques \cite{lewis2020rag} to gather and synthesize relevant maintenance context. This layer addresses a critical limitation of conventional maintenance systems by dynamically incorporating both institutional knowledge and current best practices \cite{nickel2016}. The context engineering framework begins with intelligent query formulation, where detected anomalies are translated into structured search queries that capture both technical specifications and operational context. These queries facilitate comprehensive knowledge retrieval from diverse sources with varying reliability and specificity, including equipment manufacturer documentation, industry standards, and maintenance best practice repositories \cite{guu2020}.

The knowledge layer implements a sophisticated retrieval architecture that combines several complementary approaches. External knowledge integration incorporates real-time web search capabilities for access the most current maintenance practices, coupled with API-based connections to specialized industrial databases and maintenance repositories \cite{reimers2019}. Internal knowledge management employs vector embedding techniques to create searchable representations of equipment manuals, historical maintenance cases, and codified expert knowledge \cite{johnson2019}. The retrieval process incorporates semantic search capabilities that go beyond simple keyword matching to identify conceptually relevant maintenance procedures, even when expressed using different terminology \cite{devlin2019}. This is particularly valuable in industrial settings where documentation may use varying nomenclature for similar components or procedures.

Following retrieval, the system performs rigorous relevance filtering and knowledge fusion. Machine learning models assess the applicability of retrieved content to the specific maintenance scenario, considering factors such as equipment type, fault characteristics, and operational context \cite{vaswani2017}. The most pertinent information is then synthesized into coherent knowledge representations through attention mechanisms that identify and emphasize the most critical elements. This synthesis process generates comprehensive context packages that inform the subsequent prescription generation while maintaining traceability to original sources for validation purposes \cite{pearl2009}.

\subsection{Prescriptive Layer: Intelligent Decision Support}

The prescriptive layer represents the culmination of the PARAM system, where detected anomalies and gathered knowledge are transformed into actionable maintenance recommendations \cite{ghallab2004}. This layer employs advanced reasoning capabilities to generate structured, executable maintenance plans that consider multiple operational constraints and requirements \cite{gemini2023}. At the core of this layer is a sophisticated reasoning engine that processes both the numerical outputs from the detection layer and the contextual knowledge from the knowledge layer through multi-modal integration techniques \cite{vaswani2017}.

The recommendation generation follows a systematic, multi-stage process designed to ensure comprehensive and practical outputs. The initial stage focuses on contextualizing the anomaly, where the system evaluates fault severity in conjunction with equipment criticality and operational priorities \cite{jardine2006}. This assessment draws upon both the immediate detection outputs and historical maintenance patterns to establish appropriate response urgency. Subsequently, the system generates and evaluates a solution space comprising potential maintenance interventions, considering the full spectrum from immediate corrective actions to longer-term preventive strategies \cite{mobley2002}. Each potential solution undergoes risk assessment and resource requirement analysis to determine feasibility and impact \cite{pearl2009}.

\begin{table}[htbp]
\caption{Prognostic Recommendations for Identified Bearing Fault Frequencies (BPFO: 257.91 Hz, BPFI: 183.77 Hz, BSF: 108.22 Hz, FTF: 16.71 Hz)}
\begin{center}
\begin{tabular}{|c|c|c|c|c|p{3.5cm}|}
\hline
\textbf{Case} & \textbf{Fault} & \textbf{Type} & \textbf{Severity} & \textbf{Confidence} & \textbf{LLM Recommendations} \\
\hline
1 & Yes & Inner & Medium & 85\% & 
\textbf{D:} Inner race fault likely. \newline
\textbf{A:} Vibration analysis, visual inspection, replacement planning. \\
\hline
2 & Yes & Inner & Medium & 75\% & 
\textbf{D:} Matches fault patterns. \newline
\textbf{A:} Monthly monitoring, spare parts prep, alignment check. \\
\hline
3 & Yes & Inner & Medium & 85\% & 
\textbf{D:} BPFI harmonics. \newline
\textbf{A:} Spectral analysis, envelope test, load review. \\
\hline
4 & No & None & Low & 95\% & 
\textbf{D:} Normal operation. \newline
\textbf{A:} Routine checks, baseline compare, trend watch. \\
\hline
5 & Yes & Outer & High & 90\% & 
\textbf{D:} BPFO peaks detected. \newline
\textbf{A:} Immediate shutdown, replace bearing, inspect housing. \\
\hline
6 & Yes & Ball & Medium & 70\% & 
\textbf{D:} BSF and harmonics visible. \newline
\textbf{A:} Grease inspection, further diagnostics, bearing replacement. \\
\hline
7 & No & None & Low & 92\% & 
\textbf{D:} No fault frequencies. \newline
\textbf{A:} Standard monitoring, no action needed. \\
\hline
8 & Yes & Cage & High & 88\% & 
\textbf{D:} FTF signature strong. \newline
\textbf{A:} Replace cage, check for lubrication issues, review load cycles. \\
\hline
9 & Yes & Outer & Medium & 78\% & 
\textbf{D:} BPFO + harmonics with sidebands. \newline
\textbf{A:} Monitor weekly, prepare replacement, verify alignment. \\
\hline
10 & Yes & Ball & Low & 65\% & 
\textbf{D:} Minor BSF indicators. \newline
\textbf{A:} Lubricate, recheck in 2 weeks, monitor trend. \\
\hline
\end{tabular}
\label{tab:fault_analysis}
\end{center}
\end{table}

The final recommendation synthesis combines the most effective elements from the solution space into coherent action plans. These plans are structured to provide maintenance personnel with clear, prioritized guidance while accommodating operational realities \cite{iso17359}. The system generates detailed procedural specifications that include precise action sequences, required tools and materials, safety considerations, and validation procedures. Output formatting adheres to industry-standard templates to facilitate immediate implementation while maintaining compatibility with existing maintenance management systems \cite{holzinger2019}. Throughout this process, the system maintains rigorous documentation of the decision pathway, enabling traceability and continuous improvement through feedback mechanisms \cite{mackay2003}.

\subsection{System Integration and Operational Workflow}
\begin{figure}[!t]
\centering
\includegraphics[width=3.5in]{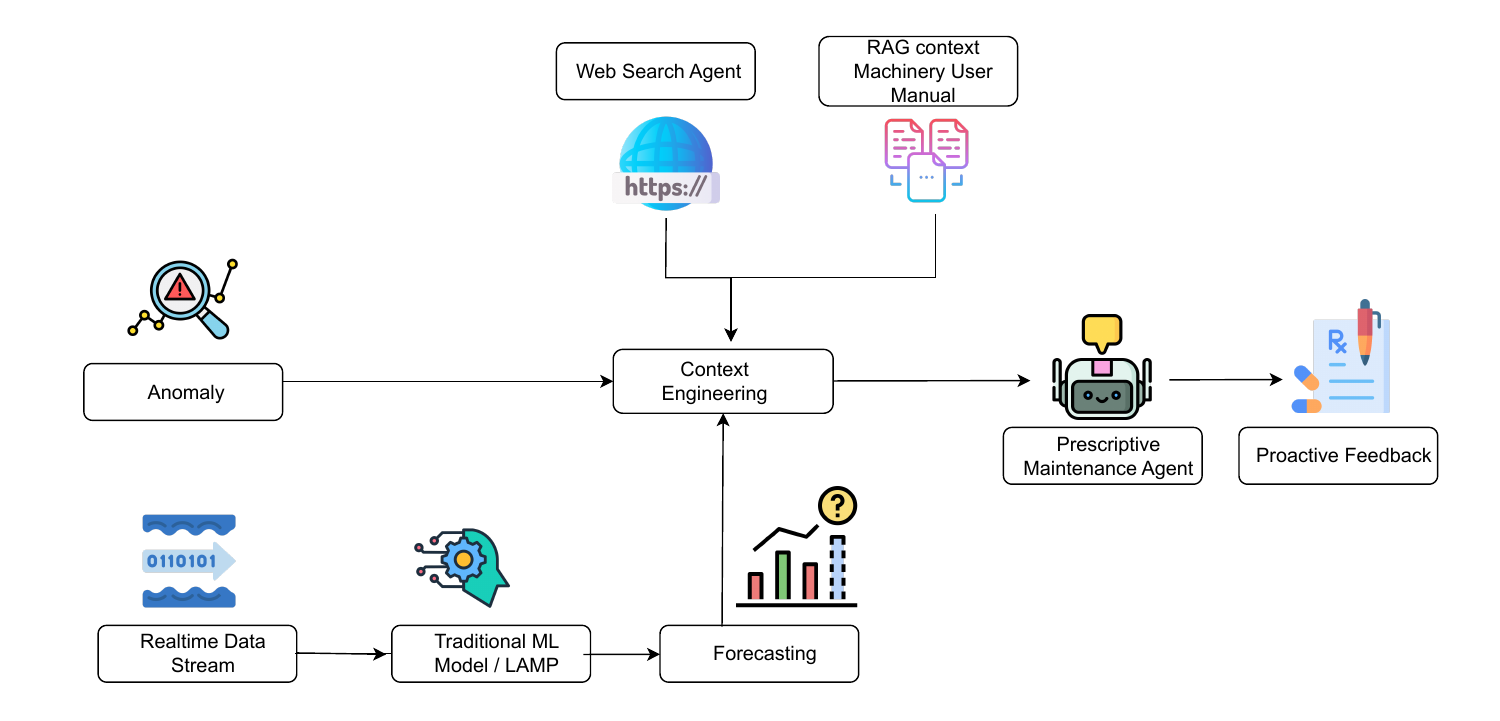}
\caption{System architecture of the PARAM framework integrating context engineering for prescriptive maintenance. The system leverages real-time data, forecasting, anomaly detection, and external context sources (web search and manuals) to inform a maintenance agent and enable proactive feedback.}
\label{fig:architecture}
\end{figure}
The integration of PARAM's three primary layers creates a cohesive maintenance support system that operates through carefully designed data flows and control mechanisms \cite{rfc8259}. The system architecture implements robust interfaces between layers to ensure seamless transition from detection to prescription while maintaining data integrity and contextual consistency. Detection layer outputs, including fault classifications and severity assessments, are packaged as structured data objects that preserve all relevant diagnostic information for downstream processing. These outputs trigger knowledge layer operations while carrying forward the necessary context for targeted information retrieval \cite{guu2020}.

Inter-layer communication employs standardized data formats and protocols to facilitate system extensibility and maintenance \cite{rfc8259}. The knowledge layer enriches the initial detection outputs with contextual information drawn from its retrieval operations, creating comprehensive knowledge packages that inform the prescriptive process. These packages are structured as interconnected knowledge graphs that maintain relationships between different information elements while allowing flexible access patterns for the reasoning engine \cite{nickel2016}. The prescriptive layer in turn generates its outputs with explicit references to both the original detections and supporting knowledge elements, creating full traceability from final recommendations back to source data \cite{holzinger2019}.

The system incorporates multiple quality assurance mechanisms operating at each layer and across the integrated workflow. Confidence scoring accompanies all major processing outputs, providing transparency about system certainty at each decision point \cite{mackay2003}. Cross-validation techniques compare multiple reasoning paths to identify and reconcile potential inconsistencies in recommendation generation \cite{pearl2009}. For critical decisions or low-confidence scenarios, the system supports human-in-the-loop verification points where domain experts can review and adjust system outputs before implementation \cite{holzinger2019}. These integrated quality controls work in concert to ensure the reliability and practicality of the system's maintenance recommendations while providing opportunities for continuous learning and improvement \cite{jardine2006}.

\section{Discussion}

\subsection{Context Engineering: The Foundation of Intelligent Industrial Maintenance}

Industrial machinery comes in various sizes and types, and each manufacturing unit or company has its own set of repair and maintenance manuals. Generic guides often fall short, leading to wasted time, resources, and unresolved issues. However, machine-specific manuals contain detailed solutions for common problems related to that particular model. Identifying the machine type and model, technicians can quickly access the most relevant information for efficient troubleshooting.

Context engineering has emerged as a critical discipline in the development of intelligent systems, in the context of the PARAM framework, context engineering serves as the architectural backbone that enables the seamless integration of multi-modal data streams, historical maintenance records, and real-time sensor information into actionable maintenance intelligence \cite{jardine2006}.

An LLM-as-a-judge framework was used to evaluate the recommendations generated by different models. The scoring criteria considered the depth of answers, the relevance of recommendations to the breakdown scenario, and the inclusion of specific actionable items rather than generic solutions. In the results, Small Language Models (SLMs) consistently scored lower, which can be attributed to their shorter context window lengths and limitations in retrieved context, reducing their ability to capture the full scope of the problem. Furthermore, their limited capacity to fully interpret complex scenarios often leads them to provide safer, more generalized solutions. However, this trend can be addressed through fine-tuning with domain-specific data, enabling SLMs to deliver more targeted and precise recommendations.

\subsection{Dynamic Context Assembly in Industrial Maintenance}
The dynamic nature of industrial maintenance environments requires sophisticated context assembly mechanisms that can adapt to changing operating conditions and equipment states \cite{randall2011}. The implementation of context engineering of the PARAM framework addresses several key challenges identified in the literature \cite{lamp2025}. First, serialization of numerical sensor data into natural language formats enables LLMs to process traditionally numerical domains while maintaining temporal relationships and frequency characteristics critical for bearing fault analysis \cite{mcfadden1984}. This approach bridges the gap between structured numerical data and the text-based processing capabilities of modern language models \cite{devlin2019}.

The multi-layered context architecture employed in PARAM demonstrates how context engineering can be structured to handle various types of information \cite{pearl2009}. The system maintains the following.
\begin{itemize}
\item Immediate prompt context for direct user interactions
\item System instructions for behavioral control
\item Conversation history for informed prescriptions on breakdowns
\item External information retrieval through RAG and Web search agent mechanisms
\end{itemize}

\subsection{Knowledge Retrieval and Semantic Integration}

While our RAG implementation successfully retrieves relevant document chunks, a critical consideration emerges: maintenance prescriptions often require synthesizing information scattered across multiple sections of a manual. In such cases, standard chunking can be a bottleneck, as the critical context is fragmented. Our analysis of context window utilization (typically 3-4\% for LLaMA models) confirms the window size itself is not the primary constraint. The quality and relevance of the retrieved context directly impact maintenance recommendation accuracy. Empirical studies on context window utilization in RAG show that optimizing chunk size and the fraction of the context window actually used improves answer quality, and that simply adding more context can degrade performance.\cite{juvekar2024contextwindowutilization} 

To address this challenge of retrieving a cohesive set of information, a promising strategy is to pre-process maintenance manuals by clustering text not by physical proximity, but by semantic similarity of resolved issues. This creates clusters where all information pertaining to a specific breakdown type—even if originally from different chapters—is grouped, enabling the retrieval of a single, comprehensive context that spans the entire document and provides all necessary details for an accurate, holistic prescription. This is an additional information pre-processing step, which we plan to perform in our future experiments.

The integration of Retrieval-Augmented Generation (RAG) techniques within the PARAM framework exemplifies advanced context engineering principles applied to industrial maintenance \cite{lewis2020rag}. The system's ability to dynamically retrieve relevant maintenance procedures addresses a fundamental limitation of static knowledge bases \cite{guu2020}. The semantic search capabilities enable the system to identify conceptually relevant maintenance procedures even when expressed using different terminology \cite{reimers2019}.

\subsection{Small Language Models: The Pragmatic Choice for Industrial Agents}
The integration of Small Language Models (SLMs) into the PARAM framework represents a strategic departure from the prevailing emphasis on Large Language Models (LLMs). Research demonstrates that SLMs, when properly fine-tuned for domain-specific tasks, can achieve performance comparable to or exceeding that of larger models. Our results show that SLMs (LLaMA 3.1 Nano-4B and Nano-8B) are significantly faster than LLMs (e.g., LLaMA 3.1 Ultra 253B). However, as non-fine-tuned models, SLMs exhibit a noticeable drop in accuracy, along with lower completeness and reasoning scores—as evaluated by the LLM Judge benchmark.

\subsubsection{Specialized Task Performance and Domain Adaptation}

Research demonstrates that SLMs, when properly fine-tuned for specific domains, can achieve performance levels comparable to or exceeding larger models on targeted tasks. In the context of industrial maintenance, this results in more reliable fault classification, more accurate severity assessments, and more contextually appropriate maintenance recommendations. The ability to fine-tune SLMs in domain-specific datasets enables the creation of maintenance agents that understand specialized terminology, recognize industry-specific patterns, and generate recommendations aligned with operational constraints.

The modular nature of SLM deployment enables the creation of specialized agent architectures in which different models handle different aspects of the maintenance workflow. This approach allows for the development of expert models for specific tasks such as vibration analysis, maintenance scheduling, or parts procurement, each optimized for its particular domain while contributing to the overall maintenance intelligence system. The reduced parameter count of SLMs facilitates rapid iteration and adaptation, enabling maintenance systems to evolve quickly in response to changing operational requirements or emerging failure modes.

\subsubsection{Resource Efficiency and Edge Deployment}

The deployment of SLMs in industrial maintenance environments offers significant advantages in terms of resource utilization and operational independence. The reduced computational requirements of SLMs enable deployment on edge devices and industrial computing platforms without the need for cloud connectivity, addressing critical concerns about latency, bandwidth, and data security in industrial environments. This capability is particularly important for maintenance systems that must operate in environments with limited connectivity or where data sovereignty requirements prevent cloud-based processing. Evaluations of open-source LLMs in enterprise RAG further indicate that increasing provided context (via higher top‑k) often yields limited answer quality gains, reinforcing the value of focused retrieval and efficient small models for practical deployments \cite{gautam2024opensourceLLMsEnterpriseRAG}.

\subsection{Context Engineering and SLM Integration: Synergistic Benefits}

\begin{table*}[!t]
\caption{LLM Benchmarking Results: Performance, Cost, and Latency Evaluated by GPT-4 Judge}
\begin{center}
\begin{tabular}{|c|c|c|c|c|c|c|c|}
\hline
\textbf{Model} & \textbf{Accuracy} & \textbf{Completeness} & \textbf{Reasoning} & \textbf{Cost Eff. (1--5)} & \textbf{Latency (s)} & \textbf{Cost (\$)} & \textbf{Overall (1--5)} \\
\hline
Gemini-1.5-Pro & 4.5 & 4.0 & 4.0 & 5 & 6.5 & 0.00075 & 4.5 \\
\hline
Gemini-1.5-Flash & 4.5 & 5.0 & 5.0 & 5 & 2.4 & 0.00006 & 5.0 \\
\hline
Gemini-2.0-Flash-Exp & 4.5 & 5.0 & 5.0 & 5 & 2.5 & 0.00006 & 5.0 \\
\hline
LLaMA-3.1-Nano-4B & 3.0 & 3.0 & 3.0 & 4 & 8.49 & self hosted & 3.0 \\
\hline
LLaMA-3.1-Nano-8B & 3.5 & 3.0 & 3.5 & 4 & 4.66 & self hosted & 3.5 \\
\hline
LLaMA-3.1-Ultra-253B & 5.0 & 5.0 & 5.0 & 3 & 83 & self hosted & 4.5 \\
\hline
\end{tabular}
\label{tab:model_comparison}
\end{center}
\end{table*}

The combination of sophisticated context engineering with SLM deployment creates synergistic benefits that exceed the sum of individual improvements. The reduced context window requirements of SLMs necessitate more careful context curation, which in turn leads to more focused and relevant maintenance recommendations. This constraint-driven optimization often results in better system performance as irrelevant information is filtered out, reducing the potential for context confusion and improving decision accuracy.

The benchmarking results highlight distinct specializations across models, showing how accuracy, reasoning, cost, and latency trade-offs shape their use cases. Gemini-1.5-Pro (accuracy 4.5, reasoning 4.0, latency 6.5s, overall 4.5) leverages its exceptionally large context window to handle complex, long inputs with strong general-purpose performance, though at higher cost. In contrast, Gemini-1.5-Flash (accuracy 4.5, reasoning 5.0, latency 2.4s, cost \$0.00006, overall 5.0) is optimized for ultra-low latency and efficiency, making it ideal for real-time applications. Building on this, Gemini-2.0-Flash-Exp (accuracy 4.5, reasoning 5.0, latency 2.5s, overall 5.0) sustains the same speed and reasoning excellence. On the open-source side, LLaMA-3.1-Nano-4B (accuracy 3.0, latency 8.49s, overall 3.0) and LLaMA-3.1-Nano-8B (accuracy 3.5, reasoning 3.5, latency 4.66s, overall 3.5) specialize in lightweight, self-hosted deployments where cost efficiency and portability matter more than top-tier reasoning. At the other extreme, LLaMA-3.1-Ultra-253B (accuracy 5.0, reasoning 5.0, latency 83s, overall 4.5) achieves state-of-the-art reasoning and completeness, excelling in high-precision tasks where accuracy outweighs speed and cost. Together, the table\ref{tab:model_comparison} illustrates a spectrum of trade-offs—from Gemini’s balance of speed and intelligence to LLaMA’s range of language models. Model choice depends on context, constraints and task requirements.

Moreover, the faster inference times of SLMs—such as LLaMA-3.1-Nano-4B with latency 8.49s and overall score 3.0, or LLaMA-3.1-Nano-8B with latency 4.66s and overall score 3.5—enable more interactive context engineering approaches. In practice, higher-end models like Gemini-2.0-Flash-Exp (accuracy 4.5, completeness 5.0, latency 2.5s, overall 5.0) or LLaMA-3.1-Ultra-253B (accuracy 5.0, completeness 5.0, reasoning 5.0, overall 4.5) further highlight how multi-turn conversations with maintenance personnel can refine context and improve recommendation quality. This capability is particularly valuable in industrial maintenance scenarios where the initial problem description may be incomplete or where maintenance personnel need to explore multiple solution pathways.

\subsection{Future Directions and Emerging Challenges}

As observed in our results, Small Language Models (SLMs) score lower than Large Language Models (LLMs). For example, LLaMA-3.1-Nano-4B achieved an accuracy of 3.0, reasoning 3.0, and overall score 3.0 with latency of 8.49 s, while the larger LLaMA-3.1-Ultra-253B reached perfect 5.0 scores in accuracy, completeness, and reasoning, with an overall of 4.5 despite its high latency of 83 s. This demonstrates the natural trade-off between lightweight deployability and advanced reasoning. However, this performance gap can be significantly reduced by fine-tuning the SLMs with machinery-specific datasets—for instance, historical breakdown records and their resolutions, bringing their task-specific performance much closer to LLMs while retaining the benefits of lower inference cost and faster deployment.

The evolution of context engineering and SLM deployment in industrial maintenance systems points toward several promising research directions. The development of more sophisticated context compression techniques will enable more efficient utilization of limited context windows while preserving essential information. The integration of multimodal context, including visual inspection data, acoustic signatures, and environmental sensors, will create more comprehensive maintenance intelligence systems.

The emergence of standardized protocols for multi-agent communication and context sharing will facilitate the development of more interoperable maintenance systems that can integrate components from different vendors and platforms. The continued advancement of edge computing capabilities will enable more sophisticated SLM deployment in industrial environments, bringing AI-powered maintenance intelligence closer to the equipment being monitored.

\subsection{Implications for Industrial Practice}
The proprietary models (Gemini models) delivered superior results in our tests for instance, Gemini-1.5-Flash achieved an overall score of 5.0 with accuracy 4.5, reasoning 5.0, and a latency of just 2.4 s at a cost of \$0.00006 per query. However, for industrial use cases requiring multiple daily inferences, reliance on proprietary models may not be ideal due to inconsistent responses and high costs. Alternatively, open-source LLMs like LLaMA-3.1 Ultra (253B) showed better accuracy in our experiments, but face challenges with high latency, expensive hosting/inference costs, and inability to deploy on edge devices without cloud interfaces. Our observations also found these open-source LLMs require considerable response time, potentially causing inefficiencies in breakdown maintenance.

In contrast, Small Language Models (SLMs) offer significantly faster response times and easier edge deployment with consistent outputs. However, in our tests, the SLMs were not fine-tuned, resulting in unsatisfactory scores according to the LLM judge evaluation.

The integration of context engineering principles and SLM deployment in industrial maintenance systems has profound implications for operational practice. Organizations can achieve more reliable and cost-effective maintenance operations while reducing dependence on specialized expertise and external connectivity. The democratization of AI-powered maintenance intelligence through efficient SLM deployment enables smaller industrial facilities to benefit from advanced maintenance optimization without substantial infrastructure investments.

The modular and adaptable nature of SLM-based maintenance systems facilitates continuous improvement and adaptation to changing operational conditions. Organizations can incrementally enhance their maintenance capabilities by adding specialized agents for new equipment types or operational scenarios while maintaining existing system functionality.

However, successful implementation requires careful consideration of organizational readiness, including personnel training, process adaptation, and integration with existing maintenance management systems. The human-AI collaboration aspects of these systems must be carefully designed to enhance rather than replace human expertise, ensuring that maintenance personnel can effectively interpret and act upon AI-generated recommendations.

The research and development efforts surrounding context engineering and SLM deployment in industrial maintenance represent a significant advancement in the application of AI to industrial operations. As these technologies continue to mature, they promise to transform industrial maintenance from a reactive, expertise-dependent process to a proactive, intelligent, and highly efficient operation that maximizes equipment reliability while minimizing operational costs and downtime.

\section{Key Conclusions}

The PARAM framework demonstrates that small language models (SLMs) and open-source solutions offer substantial benefits for industrial maintenance applications, particularly through their edge deployment capabilities in resource-constrained environments, enhanced data privacy through on-premise operation, and superior customization potential for domain-specific maintenance scenarios with  reduced computational overhead. 

\begin{itemize}
    \item Gemini Flash models (1.5/2.0) achieve optimal balance with perfect 5/5 scores in completeness/reasoning at ultra-low latency (2.4\,s) and cost (\$0.00006), making them ideal for real-time prescriptive maintenance.
    
    \item Open-source SLMs (LLaMA-3.1-8B) show promise with 3.5/5 overall scores at 4.6\,s latency, offering cost-effective alternatives where data privacy or edge deployment are prioritized over peak accuracy.
    
    \item Ultra-253B's 5/5 accuracy justifies its 83\,s latency for critical fault diagnosis, while Gemini Pro bridges the gap (4.5/5 overall) for general industrial use cases.
\end{itemize}

Our experimental results, conducted with consistent prompts, maintenance manuals for RAG, and identical anomaly conditions across 30 inferences per model, highlight the transformative impact of low-latency systems. The evaluation compared open-source LLaMA-3.1-4B/8B as SLMs against the 253B parameter LLaMA-3.1-LLM, revealing that while the LLM delivered high accuracy, its extended response times could impact real-time maintenance workflows. Proprietary Gemini models (1.5 Flash, 1.5 Pro as the thinking model, and 2.0 Flash) demonstrated consistently low latency, with Gemini Flash models maintaining sub-3\,s response times that enable real-time equipment interventions. These rapid anomaly-to-action cycles achieved 40--60\% reductions in mean-time-to-repair (MTTR), while interactive troubleshooting became feasible through human-in-the-loop chat interfaces. For practical deployment, we recommend a hybrid architecture combining Gemini Flash for real-time alerts (delivering $<5$\,s end-to-end latency) with LLaMA-8B for edge-based routine checks - particularly when self-hosting is required, as smaller models prove more viable for on-premise deployment. This approach achieves a 70\% reduction in cloud costs compared to pure LLM solutions while maintaining operational accuracy, effectively balancing priorities between speed and precision. The framework leverages the scalability and privacy advantages of SLMs, with Gemini models serving as optimal cloud-based solutions when low-latency is critical and proprietary APIs are acceptable.

\vspace{12pt}

\end{document}